\documentclass[runningheads]{llncs}
\usepackage{graphicx}
\usepackage{booktabs}
\usepackage{ragged2e}
\usepackage{multirow}
\usepackage{amsmath}

\usepackage{xspace}
\usepackage{graphbox}
\usepackage[
backend=biber,
style=numeric,
sorting=ynt
]{biblatex}
\usepackage{rotating}
\usepackage{diagbox}
\usepackage{tikz}

\usepackage{makecell}
\usepackage{array}
\usepackage{vcell}
\usepackage{multirow}
\usepackage[inline]{enumitem}

\usepackage{orcidlink}
\usepackage{cleveref}

\newcommand{\MIDVHOLO}{MIDV-Holo\xspace}
\newcommand{\MIDVHOLOREF}{MIDV-Holo~\cite{koliaskina_midv-holo_2023}\xspace}

\newcommand{\random}{uniformly selected\xspace}

\newcommand{\PUBLICURL}{\url{https://github.com/EPITAResearchLab/pouliquen.24.icdar}}

\begin{document}
\title{Weakly Supervised Training for\\Hologram Verification in Identity Documents} %
\author{%
Glen Pouliquen\inst{1,2}\orcidlink{0009-0002-5231-2228} \and
Guillaume Chiron\inst{1}\orcidlink{0009-0004-3665-4900} \and
Joseph Chazalon\inst{2}\orcidlink{0000-0002-3757-074X} \and\\
Thierry G{\'e}raud\inst{2}\orcidlink{0000-0002-0380-7948} \and
Ahmad Montaser Awal\inst{1}\orcidlink{0000-0002-0479-6312}%
}
\authorrunning{Pouliquen et al.}
\institute{%
IDnow AI \& ML Center of Excellence, Cesson-Sévigné, France \\
\email{name.surname@idnow.io}%
\and
EPITA Research Lab. (LRE), Le Kremlin-Bicêtre, France \\
\email{name.surname@epita.fr}
}
\maketitle
\begin{abstract}
We propose a method to remotely verify the authenticity of Optically Variable Devices (OVDs), often referred to as ``holograms'', in identity documents.
Our method processes video clips captured with smartphones under common lighting conditions, and is evaluated on two public datasets: MIDV-HOLO and MIDV-2020.
Thanks to a weakly-supervised training, we optimize a feature extraction and decision pipeline which achieves a new leading performance on MIDV-HOLO,
while maintaining a high recall on documents from MIDV-2020 used as attack samples.
It is also the first method, to date, to effectively address the photo replacement attack task, and can be trained on either genuine samples, attack samples, or both for increased performance.
By enabling to verify OVD shapes and dynamics with very little supervision, this work opens the way towards the use of massive amounts of unlabeled data to build robust remote identity document verification systems on commodity smartphones. Code is available at \PUBLICURL.

\keywords{Know Your Consumer (KYC) \and Identity Documents \and Hologram Verification \and Weakly Supervised Learning \and Contrastive Loss}
\end{abstract}

\begin{figure}[!bt]
\centering
\includegraphics[width=\textwidth]{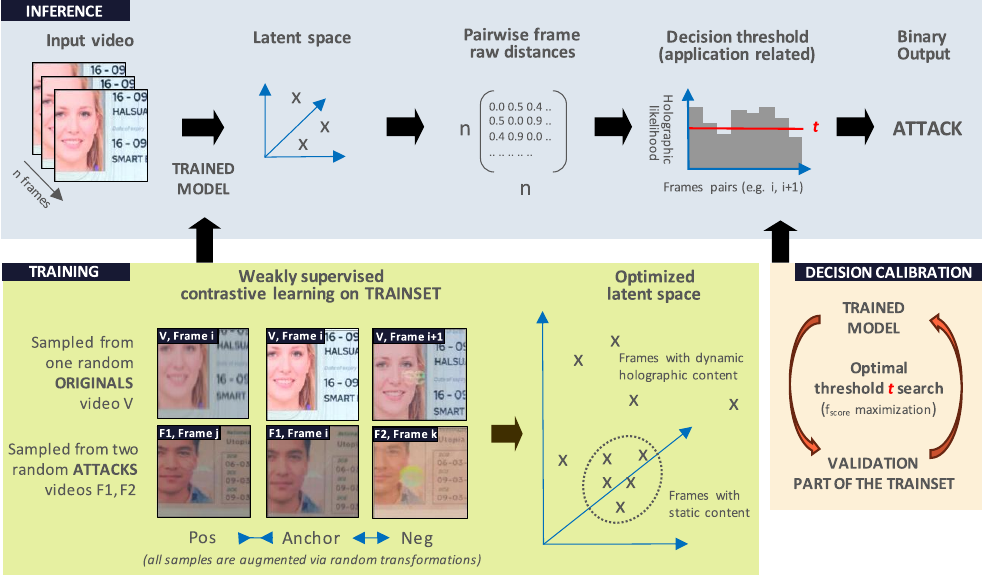}
\caption{Proposed approach overview, involving 1) the weakly supervised \textsc{training} with a specific data selection strategy over the trainset; 2) the \textsc{inference} pipeline extracting optimized features used afterward to compute the final ``Original/Attack'' decision based on a thresholding of pairwise distances. The \textsc{decision} illustrates how the threshold is calibrated over the validation part of the train set.}
\label{fig:weaklysupervisedpipeline}
\end{figure}
\section{Introduction}
\begin{figure}[tb]
\resizebox{\textwidth}{!}{
\includegraphics[]{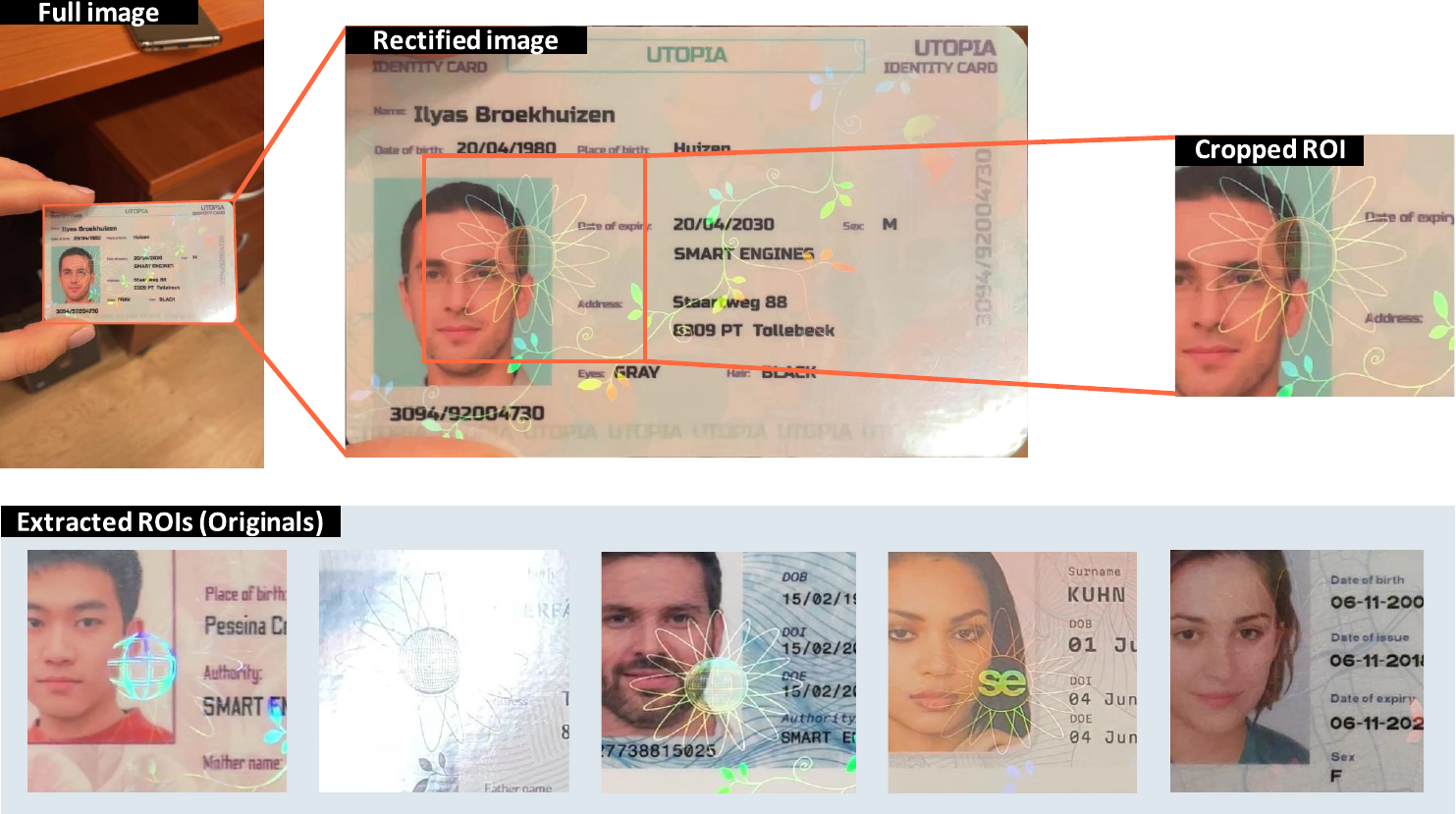}}
\caption{In MIDV Holo dataset, documents are captured in different places involving various backgrounds and lightning conditions (left). Document quads are annotated on all images allowing rectifications (center). Additionally, we propose to define a region of interest containing part of the face and the holograms in charge of securing it (right). Extracted Regions of Interest (ROIs) from sampled labeled as "Originals" (below) contain more or less visible holographic content. \emph{Identities} (names and faces) are synthetic.}
\label{fig:midv-holo-demo}
\end{figure}

Often called KYC (Know Your Customer), remotely verifying the authenticity of identity documents is a critical point for building online trust. This is an increasingly regulated process which relies on identity documents, among other proofs, to establish the link between an online identity and a real state-backed one. This linking usually requires checking that the document is original and was not altered. The photography of the bearer is of paramount importance here to ensure that the user of a remote system is the intended one.
Optically variable devices (OVDs), commonly referred to as ``holograms'' and illustrated in \Cref{fig:midv-holo-demo}, are powerful tools to secure physical documents in line with the recommendation of the EU council \cite{prado}, among others.
Built using elaborated and undisclosed optical techniques, these devices exhibit rich visual behaviors when viewing and/or illuminating conditions (angle, light color, etc.) change.
They are embedded in a wide spectrum of sensitive elements, i.e. not only identity documents but also banknotes or tamper-proof labels (e.g. for medical drugs), and contribute to ensure:
\begin{enumerate*}
    \item \textbf{Integrity:} they cannot be removed without altering their properties, making tampering very challenging;
    \item \textbf{Authenticity:} they are very difficult to forge, making the creation of fraudulent documents equally hard.
\end{enumerate*}

Despite the widespread use of holograms, automating their remote verification in the context of an automatic enrollment, whether it is to open an account in an online bank or to contract a loan, poses many challenges.
Indeed, such visual objects were primarily designed for manual inspection, sometimes using special tools like magnifiers or dedicated light sources.
As a result, automated remote validation is limited in many aspects: acquisition is often performed using commodity smartphones under uncontrolled ambient light to capture macroscopic and visible patterns, while following simple interactive scenarios.
Nevertheless, verifying holographic devices from a video is possible to some extent, and many recent works and datasets contributed to this effort.

While, in the general forgery detection literature, several approaches try to detect falsification clues \cite{nirkin2021deepfake}, others follow the opposite (yet complementary) direction of checking whether clues of authenticity and integrity are present \cite{Lin2019FaceLD}. 
This work contributes to the latter:
we propose a method to control the presence of some holographic content at specific positions of a document (e.g. photography area), and address the problem of photography replacement, which was introduced in \MIDVHOLOREF but not yet addressed (to our knowledge).
After a detailed review of related approaches and datasets (\Cref{sec:relworks}),
we introduce our key contribution: a new method to detect and validate holographic content, whose feature extraction is trained in a weakly-supervised fashion (i.e., not requiring a precise labeling of each video frame with the particular visual appearance of a hologram), and which outperforms the original approach on public datasets 
(\Cref{sec:proposedapproach}).
For practical reasons, we also propose an updated experimental protocol which specifies, among others, training, validation and test sets for the \MIDVHOLO dataset, as well as a public, open-source reimplementation of the approach proposed in the original \MIDVHOLO publication~\cite{koliaskina_midv-holo_2023}, with systematic optimization of the calibration of the decision function (\Cref{sec:midvproper}).
Our approach (illustrated in \Cref{fig:weaklysupervisedpipeline}) is carefully evaluated on several public datasets, over several runs, and an ablation study is conducted to challenge the benefits of every aspect of our method (\Cref{sec:experiments}).
The code to reproduce our results is publicly available at \PUBLICURL.

\section{Related works}
\label{sec:relworks}

Optically Variable Devices (OVDs) are often built using polarized inks or diffraction grating ---~a network of microscopic reflective structures engraved in the thickness of some transparent layer.
We refer the reader to the \MIDVHOLO publication~\cite{koliaskina_midv-holo_2023} for a concise introduction on the optical design of these objects.
These OVDs exhibit continuous, sometimes rapid, transitions among a virtually infinite set of visual states
when changing the relative positions of the light source(s), the camera and the document.
We can consider such space of visual appearances as a sort of manifold which we navigate by changing visualization conditions.
Such model is valuable to identify the 3 fundamental visual features an automated method can check:
\begin{enumerate}
  \item
  \textbf{Appearance Conformity:} \emph{Can a particular visual appearance (shape and color) be generated by a genuine OVD?} ---~
  This can be viewed as assessing how far a particular sample (usually an image) is from the real manifold of a given hologram, and use a distance threshold as verification criterion.
  Implementing this control enables to detect attacks with no hologram or with a different hologram shape but, if used alone, would be tricked by simple static copies of the expected hologram.
  \item
  \textbf{Appearance Coverage:} \emph{How well do a particular set of visual appearances (usually captured as a video) matches the set of possible ones?} ---~
  This can be viewed as measuring how well the samples obtained cover the real manifold of a given hologram.
  Implementing this control enables to detect attacks with static holograms but, if used without control of state conformity, would be tricked by any random holographic layer. Approaches checking only color distributions are vulnerable to this attack.
  \item
  \textbf{Transitions Validity:} \emph{Are the transitions between observed visual appearances consistent with expected ones?} ---~
  This can be viewed as checking whether samples obtained describe valid paths on the real manifold of a given hologram.
  Implementing this control enables to detect attacks with imperfect hologram imitations or rapid swapping of static holograms.  %
  The low frame rates utilized in current real remote authentication applications present a significant challenge that has not been adequately explored in the research literature.
\end{enumerate}

Early approaches like the one of Hartl et al.~\cite{hartl_efficient_2016} identified a discrete set of visual appearances to check for during a manual inspection of identity documents.
Expected visual states were acquired and validated using a robotic arm with controlled light.
While lacking automated verification, this approach proposed a practical protocol to assist a human operator during its work to validate 1. visual state conformity thanks to visual comparison, and 2. visual behavior completeness by checking every expected visual state is seen.

The work of Chapel et al.~\cite{chapel_authentication_2023} proposes a way to automate the verification of visual states conformity thanks to a learned classifier based on local binary patterns (LBP) features.
However, training this system requires to label each video frame with target class (visual state), which is both too expensive for real application in our experience, and also challenging because of the frequent ``mixing'' of visual shapes in real OVDs.
Furthermore, robustness of the static feature extractor may be a concern when background are not constant like in the case of face photos: a learned feature extractor seems necessary here.

To overcome the need for labeled frames when training a visual state classifier, some approaches relaxed the control on visual state conformity to focus instead on the validation of visual behavior completeness.
The work of Kada et al.~\cite{kada_hologram_2022} opened an interesting direction by restating the problem as a semantic segmentation problem where images of documents captured as a video are first carefully registered, then a pixel-level classification is performed to predict whether a particular region belongs to a hologram.
Such prediction is mainly based on statistics on the distribution of pixel color values. %
While the final segmentation map may be used as some visual appearance clue, this method does not check whether inter-pixel behavior is consistent, nor visual appearance conformity for a particular frame, and lacks a global decision stage.

A major step was made thanks to the work of Koliaskina et al.~\cite{koliaskina_midv-holo_2023}.
The authors not only reuse the same idea of semantic segmentation (also based on a static, handcrafted feature extraction) to produce a map of pixels which exhibit some ``holographic behavior'',
but also provide a global decision stage based on a variance threshold and a first public dataset, \MIDVHOLO, containing document with holographic contents along with several presentation attacks, as illustrated in \Cref{fig:midv-holo-demo}.
While validated on full-size documents, the approach was not tested on the particular case of the face photo region, and was not evaluated against the photo replacement attack.
Furthermore, as detailed in \Cref{sec:midvproper}, this milestone publication required us to reimplement the proposed approach and specify training, validation and test splits to conduct a fair comparison with our proposed approach.

Another family of approaches tackled the problem of learning a useful embedding space, thanks to which it may be possible to both check visual state conformity and visual behavior completeness.
A first example is the work of Soukup et al.~\cite{soukup_mobile_2017} targeting hologram verification on banknotes.
Their approach extract representations from video frames using a Convolutional Neural Network (CNN) trained with a supervised classification task.
Target classes represent different visual appearances, and were captured using a LED ring that illuminated the hologram in various directions to automate the annotation process.
Because of the changing nature of the background for some OVDs in identity documents (such as in the area of the face picture), such approach may need an important amount of training data to be applied in our context, which makes it impractical since it requires physical access to real documents. %

Finally, a last related work is the one of Ay et al.~\cite{ay_open-set_2022} which proposed to train a Generative Adversarial Network (GAN) to capture the visual properties of some hologram.
While the generative properties of such approach are very attractive, successfully training such architecture to model thin holograms on non-constant backgrounds like face pictures is a great challenge, as the network may more easily capture and generate facial feature which cover a larger proportion of the area of interest.
Furthermore, as final decision is performed using the discriminator network, a proper calibration would require attack samples.
Another limitation of using the discriminator is that, despite the rich representation learned, this approach is limited to controlling the visual state conformity of isolated frames, and cannot check visual behavior completeness as differences between visual states cannot be measured. 

Looking at these existing works, we can sketch out desirable features an automated verification method should provide, which our proposed approach tends to incorporate:
\begin{enumerate}
  \item It should be based on a learned representation which can be tied to a particular OVD, in order to be able to capture both visual appearance and behavior information, as opposed to handcrafted, static, pixel-based feature extraction techniques.
  \item Such representation should be learned in a weakly supervised way to avoid requiring manual labelling of existing frames, or physical access to a large amount of original documents and presentation attacks.
  \item The learning objective should be able to guide the training even in the presence of non-constant backgrounds and thin holographic objects, like in the case of the face picture area.
\end{enumerate}

\section{Contrastive Learning of Hologram Representation}
\label{sec:proposedapproach}

This section introduces our key contribution, summarized in \Cref{fig:weaklysupervisedpipeline}: a new method to detect and validate holographic content, whose representation (feature extraction) is trained in a weakly-supervised fashion;
i.e. it only requires a single label (``original'' or ``attack'', as per \MIDVHOLOREF terminology) for each video clip.
For this purpose, we use a particular kind of contrastive loss which enables the training to be driven by intrinsic data properties. %
This relies on certain assumptions about the video clips, %
such as their ability to capture varied perspectives of the document. It also involves various transformations to enhance the data. %
The resulting representation can be shown to effectively focus on hologram regions, %
and can be used to assess both appearance consistency and coverage in a final decision stage considering as many video frames as necessary. %

\subsection{Learning Objective}  %
\label{subsec:proposed-loss}
To avoid the need for assigning labels to every video frame of the training set, we employ a contrastive learning objective which enables us to drive model training using intrinsic data properties (described in the next subsection).  %
More specifically, we use a triplet loss~\cite{balntas_learning_2016} defined on a minibatch of $N$ elements as
\begin{equation}
\label{eq:loss}
\mathcal{L}(a, p, n) = \frac{1}{N}\sum_i^N l_i(a_i, p_i, n_i), \quad l_i(a_i, p_i, n_i) = \max\left( d(a_i, p_i) - d(a_i, n_i) + m , 0\right)
\end{equation}
where $a_i$ is the projected representation of an \emph{anchor} sample whose distance from the representation of a \emph{positive} (similar) sample $p_i$ is minimized, while the distance to the representation of a \emph{negative} (dissimilar) sample $n_i$ is maximized.
Each of these representations are computed from augmented inputs to improve training.
An extra margin term $m = 1$ is used to enforce a minimal distance to negative samples.
We use $d(x_i, y_i) = ||xi - y_i||_ 2$ as distance function,
and train using an AdamW optimizer~\cite{loshchilov_decoupled_2019} with default PyTorch parameters.

\subsection{Triplet Sample Selection Strategy}
\label{subsec:proposed-samplesel}
\begin{figure}[tb]
\includegraphics[width=1.0\textwidth]{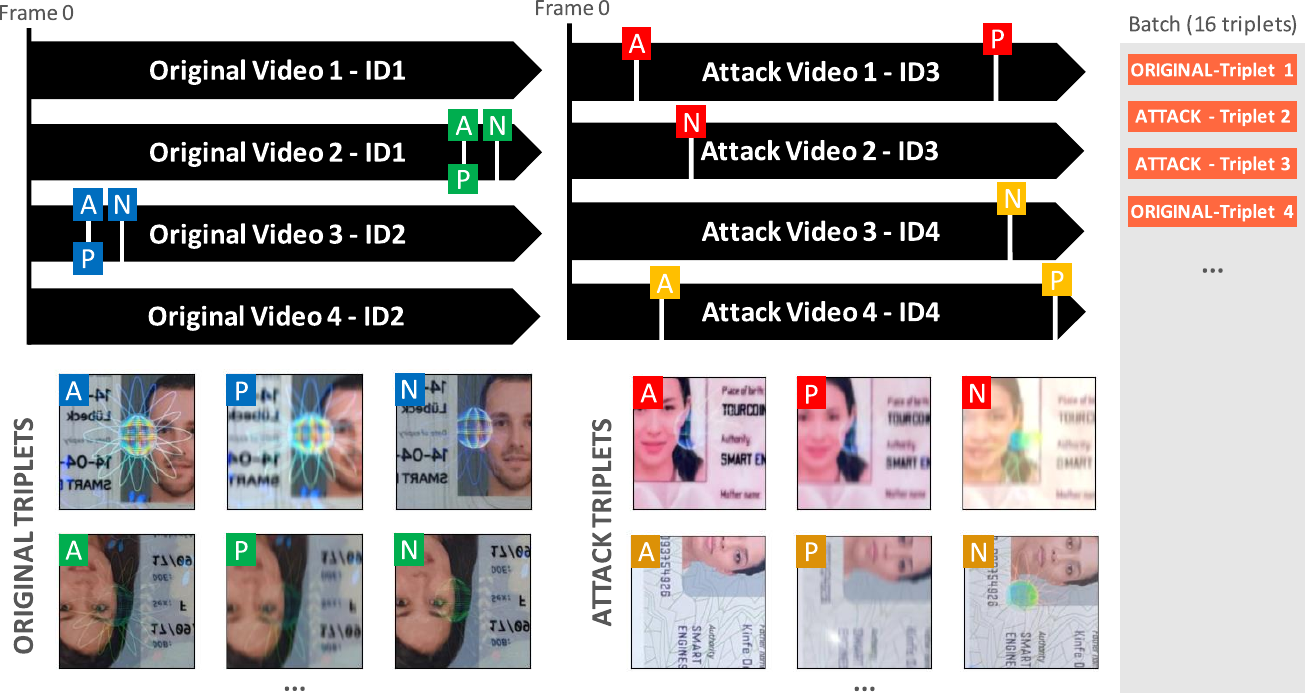}
\centering
\caption{Frame sampling strategy for building triplets [A]nchor/[P]ositive/[N]egative from \emph{original} and \emph{attack} videos. Each \emph{original} triplet is sampled from a unique \emph{original} video (A and P at $t$, N at $t+1$). Each \emph{attack} triplet is sampled from 2 different videos of a same identity with A and P, both belonging to a common video, and N belonging to a different one. All samples are transformed with \random augmentations.}
\label{fig:tripletselection}
\end{figure}

The selection of the samples which constitute the triplets is the cornerstone of our approach.
It is guided by weak labels provided at the video clip level, i.e. ``original'' or ``attack'', which exhibit different properties in the \MIDVHOLO dataset.
In the case of video clips labeled as ``originals'', we assume the visual appearance of the hologram throughout a major part of the recording.
Conversely, for video clips labeled as ``attacks'', we assume that there will be no change in its visual appearance.
This requires to remove cases of ``photo replacement'' attacks from our training set, as they exhibit the behavior of the real hologram except at the position of the replaced face picture.
These assumptions led to the following selection process, illustrated in Figure \ref{fig:tripletselection}:
\begin{itemize}
    \item \textbf{Original}: Given that the document is always moving in the videos of the dataset (at 5 frames per second), we assume that the hologram is changing. Thus, frame $t$ and frame $t+1$ are expected to contain two different visual states of the hologram. The anchor and the positive samples are generated from the same frame $t$ with different augmentations. Frame $t+1$, with augmentation, is used as the negative sample.
    \item \textbf{Attack}: In the case of an attack, we know that all the frames from a same video contain the exact same visual state of the hologram. Therefore, we take \random frames from this video as anchor and positive sample. For the negative sample, we select a \random frame from another video with the same identity.  %
\end{itemize}    
In both cases, the anchor, positive, and negative frames all represent the same identity (face picture). Consequently, the network aims to minimize the distance between the embeddings of two frames depicting the same visual state of a hologram while maximizing the difference between a frame showing the same face but with different hologram states. The assumption that the viewpoint continuously changes is generally valid in the \MIDVHOLO dataset and can be easily enforced in a real industrial scenario. This is because a document detection stage is typically required during capture to localize, classify, and rectify documents, providing indications about the camera pose relative to the document.

\subsection{Augmentations}
\label{subsec:proposed-augmentation}
To diversify anchor, positive, and negative samples (initially resized to 256 px), we apply transformations with specified probabilities:
\begin{itemize*}
    \item 
    Rotation, horizontal or vertical mirroring ($p=0.5$) applied equally to anchor, positive and negative samples.
    \item 
    Crop by a random 80\% ROI, resized to 224 px ($p=1$).
    \item 
    Gaussian blur ($p=0.4$, kernel: $3<\text{kernel}<11$, $2<\sigma<10$).
    \item 
    Color jittering ($p=0.4$, $0.7<\text{brightness}<1.3$, $0.9<\text{contrast}<1.1$, $0.95<\text{saturation}<1.05$).
\end{itemize*}
Images are then normalized to ImageNet's mean and standard deviation.

\subsection{Qualitative Validation: Feature Attribution Maps}
\label{subsec:proposed-featuremaps}
\begin{figure}[tb]
\includegraphics[width=\linewidth]{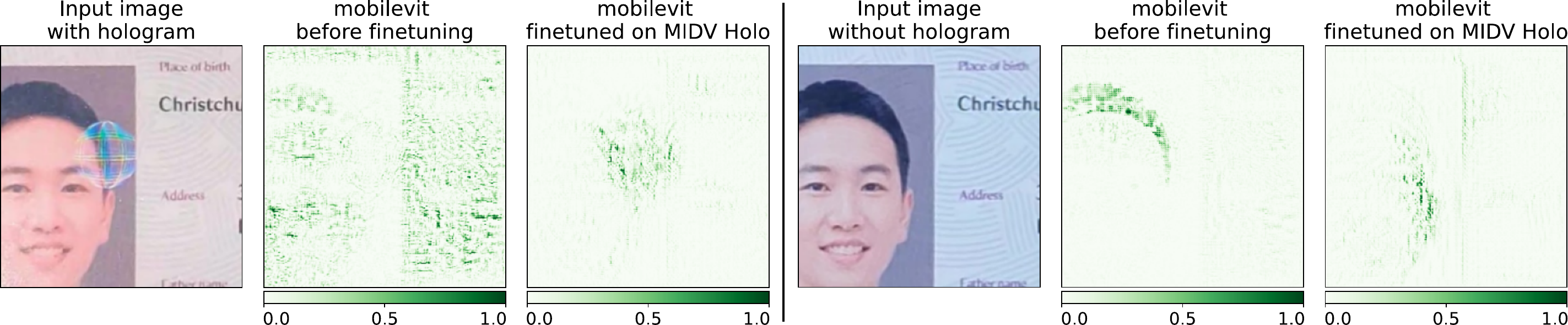}
\centering
\caption{Integrated Gradients~\cite{axiomaticattributionSundararajanTY17} visualizes a training sample, emphasizing our method's effectiveness in directing the network's attention towards the hologram. In contrast, the ImageNet-trained model lacks this focused attribution to the hologram, highlighting the significance of our training approach.}
\label{fig:axiomaticattribution}
\end{figure}

We employed the Integrated Gradients method by Sundararajan et al.~\cite{axiomaticattributionSundararajanTY17}, implemented in Captum~\cite{kokhlikyan2020captum}, to identify the focal elements of our approach. The results, illustrated in \Cref{fig:axiomaticattribution}, showcase the efficacy of our weakly supervised training method. Specifically, the model ($mobilevit_{xxs}$) trained using this approach assigns significant importance to the hologram area. This stands in stark contrast to the same network architecture trained exclusively on ImageNet, which lacks a similar level of focus on the hologram. This observation underscores the value of our training strategy in guiding the model’s attention towards pertinent features.

\subsection{Final Decision Stage}
\label{subsec:proposed-decision}
Finally, thanks to the representation produced by the feature extraction network, learned in a weakly-supervised manner as previously presented, we can extract and compare vector representations for each frame of a video clip to control. Subsequently, we compute pairwise $cosine$ similarity between the representations of the frames.
We consider two scenarios: analyzing the full video (more resource-intensive but theoretically more reliable), or adopting the incremental cumulative mode introduced in the original \MIDVHOLO publication. This last mode deems a video clip as original as soon as it meets the acceptable criterion.
By computing the mean of these differences, it becomes possible to obtain an indicator of the expected visual behaviors' coverage. We use a single threshold calibrated on the validation set to make the final decision on accepting the video clip as original or triggering an alert for a potential attack.
It is important to note that this approach does not directly inspect each individual visual appearance of the hologram. However, as the representation is trained to project non-hologram content to the same representation, such content tends to be constant
while frames containing hologram content, on the other hand, typically exhibit variability.
This results in a simultaneous control of \emph{Appearance Conformity} and \emph{Appearance Coverage}.

\section{Extensions to MIDV-Holo}
\label{sec:midvproper}
This section introduces extensions to the original \MIDVHOLO dataset and evaluation protocol that we needed to benchmark our contribution.
To compare the performance of our proposed approach with the \MIDVHOLO baseline, we re-implemented and open-sourced the latter for the ``no tracking'' mode, i.e., without the frame alignment preliminary stage.
Then, we propose an improved protocol enabling cross-validation to cope with the variance we observe in experimental results.
This requires to revise the metrics used and to ways to separate training, validation, and test sets, over several runs.

\subsection{Reproduction of the MIDV-Holo Baseline Approach}
The authors of the original \MIDVHOLO publication~\cite{koliaskina_midv-holo_2023} introduced a baseline approach for semantic segmentation of video frames, 
identifying pixels within a holographic area and computing their ratio as a proxy to verify the hologram's shape,
resulting in a binary decision:
\begin{itemize*}
\item \emph{negative}: the video clip is deemed to contain the expected hologram, considered ``original'';
\item \emph{positive}: no hologram is found, raising an alert for a potential ``attack''.
\end{itemize*}
No public implementation of this baseline approach existed, so we created a public, open-source re-implementation following the authors' guidance. We evaluated our approach using the same conditions and metrics outlined in \Cref{tab:midv-baseline-reimpl} before integrating it into our experiments.

The baseline approach, originally calibrated and tested on the entire \MIDVHOLO dataset without clear separation between calibration and test sets, relied on three parameters: $S_{\text{thresh}}$, $h_{\text{thresh}}$ and $T$.
Our implementation differs from the original in two notable aspects:
\begin{enumerate*}
\item resizing images to $1123\times709$ pixels and
\item imposing a minimum buffer of five frames before returning a result, compared to the original's theoretical requirement of two frames.
\end{enumerate*}
We recalculated Table 1 of the original paper for the ``no tracking'' mode and observed nearly identical performance in terms of ROC AUC as reported in \Cref{tab:midv-baseline-reimpl}.
Specifically, we identified the same optimal threshold configuration ($S_{\text{thresh}} = 50$ and $h_{\text{thresh}} = 0.01$) across values of the $T$ parameter.

\begin{table}[tb]
    \centering
    \caption{Reproduction of Table 1 from the original \MIDVHOLOREF publication, comparing ROC AUC values in ``no tracking'' mode between our re-implementation and the original, validating its accuracy.%
    }
    \label{tab:midv-baseline-reimpl}
    \resizebox{0.8\linewidth}{!}{%
    \begin{tabular}{|l|r|r|r|r|r|r|r|r|r|}
    \hline
    $S_\text{thresh}$ & \multicolumn{3}{r|}{30} & \multicolumn{3}{r|}{40} & \multicolumn{3}{r|}{50} \\ \hline
    $h_\text{thresh}$ & 0.01 & 0.02 & 0.03 & 0.01 & 0.02 & 0.03 & 0.01 & 0.02 & 0.03 \\ \hline
    Original \MIDVHOLOREF & 0.795 & 0.825 & 0.832 & 0.828 & 0.841 & 0.832 & \textbf{0.847} & 0.838 & 0.807 \\ \hline
    Our re-implementation & 0.838 & 0.846 & 0.844 & 0.855 & 0.844 & 0.831 & \textbf{0.857} & 0.826 & 0.790 \\ \hline
    \end{tabular}
    }
\end{table}

\subsection{Enabling Cross-Validation on \MIDVHOLO}
\label{subsec:midv-enable-crossval}
While being an important contribution with a first public dataset with ``holograms'' in identity documents,
\MIDVHOLO still contain little data: 700 video clips which can be broken down as follows:
\begin{itemize*}
    \item 2 types of documents, equally shared: identity card-like and passport-like,
    \item 10 model variants for each type, also equally shared,
    \item 5 ``identities'', i.e., fake holder for each model variant, and
    \item 3 originals and 4 presentation attacks (actual video clips) for each ``identity''.
\end{itemize*}
These presentation attacks can either contain static content, in the case of the 
``copy without holo'' (no hologram at all),
``pseudo holo copy'' (static imitation using an image editor),
and ``photo holo copy'' (photocopy of the document) attacks;
or dynamic content as in the case of the ``photo replacement'' attack where an original document is physically altered to change the face picture.
In this latter case, no hologram is visible over the face picture, but it is still present on the rest of the document.
All original document variants exhibit the \emph{same hologram}, with a small translation between identity card and passport documents.

\begin{figure}[tb]
\includegraphics[width=0.95\linewidth]{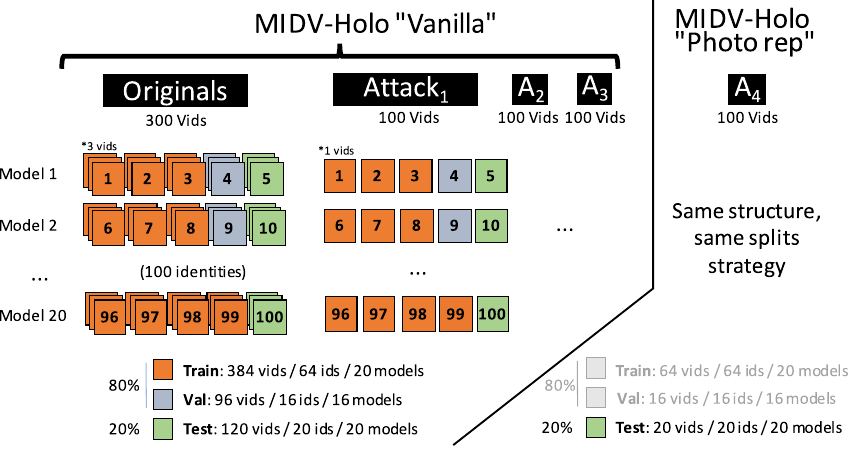}
\centering
\caption{Proposed split over the MIDV-Holo dataset (64\% train,  16\% validation and 20\% test). MIDV-Holo ``Vanilla'' refers to the part tackled in the original paper. ``Photo replacement'' attacks are exclusively used for testing in our experiments.}
\label{fig:dataset_split}
\end{figure}

Because we need to be able to compare methods trained and calibrated on this dataset and reduce variance in the experiments,
we propose to specify training, validation and test sets for 5 different splits.
The process for generating such splits is illustrated in~\Cref{fig:dataset_split},
and aims at challenging the generalization to new identities rather than the generalization to new documents.
Therefore, all identities in the train, validation, and test sets are distinct, 
while the document models (i.e. identity cards, passports) are common across the different sets.
We proceed as follows:
we stratify the dataset by document model (20 cases),
then for each document model, we select 1 out of 5 identity for testing,
the 4 remaining ones being used for training, except for 1 document every 5 items where an identity is kept for validation instead of training.
All video clips for the selected identities go into the same target subset for a given split.
This results in the following data partition: 64\% training, 16\% validation and 20\% test.
No identity can be present in two subsets for a given split.
The \emph{``photo replacement''} attack case is handled specially as we never use the corresponding samples for training or validation, and only use them for testing.

A last modification to the original protocol is that we use the $F_{score}$ (harmonic mean of Precision and Recall) as the final metric,
while \MIDVHOLO authors preferred to report Recall values for a false positive rate close to 10\%.
Reporting ROC curves computed on the test set would be possible, but would only give a hint about the expected performance in production while hiding calibration uncertainty.
In order to avoid an extra level of complexity during the training of a learned feature extractor (to favor Recall over Precision), we encourage the use of a simpler metric which provides a total ordering.

\section{Experiments and Results}
\label{sec:experiments}

This section proposes an experimental evaluation of our proposed approach (described in Section \ref{sec:proposedapproach}) compared to the \MIDVHOLO baseline~\cite{koliaskina_midv-holo_2023}, along with an ablation study.
Contrary to the \MIDVHOLO original experiment applied to the whole rectified documents (see reproduced results in \Cref{tab:results}), our experiments exclusively focus on the critical region of the document containing the face picture of the bearer.
We believe it is important to be able to leverage prior knowledge about the documents controlled, and challenged this idea by cropping registered document image to a particular region of interest for each document model, as illustrated in \Cref{fig:midv-holo-demo}. 

\subsection{Experimental Protocol}
Our experiments utilized three publicly available datasets. Initially, both our proposed method and the \MIDVHOLO baseline were trained and calibrated using the \MIDVHOLO ``Vanilla'' training and validation sets, as defined in the preceding section. Subsequently, they were assessed on three distinct test sets from the MIDV series to gauge their generalization capabilities:
\begin{itemize}
    \item \textbf{\MIDVHOLO ``Vanilla''} (120 test videos) originally introduced in~\cite{koliaskina_midv-holo_2023}, selecting only test set elements as defined in~\Cref{subsec:midv-enable-crossval}.
    \item \textbf{\MIDVHOLO ``Photo Replacement''} (20 test videos) is a specific subset of \MIDVHOLO, and represents a distinct task from the ``Vanilla'' set, as it was not addressed in the original paper. While the complete set comprises 100 videos, our experiments solely involve the 20 test videos at each run, as our approach does not entail training on this dataset.
    \item \textbf{MIDV 2020 ``Clips''} (1000 videos, test only): To assess the method's generalization to various document types, we utilized images from the ``Clips'' category of the MIDV 2020~\cite{bulatov_midv-2020_2022} dataset.
    Following document rectification, similar Regions of Interest (ROIs) were extracted as for \MIDVHOLO. As clips were sampled at 10 frames per second (fps), we dropped one frame out of two to match the frame rate of \MIDVHOLO (5 fps).
\end{itemize}
It's important to note that the \MIDVHOLO dataset features a single form of holographic layer, consistent across all 20 document models, with minor translations between identity-card-like and passport-like models. Consequently, our system effectively trains to detect and validate this specific holographic device.

We compared several variants of the approach.
For the feature extraction stage, we tested the following models, all implemented using the timm library~\cite{rw2019timm}:
$resnet_{18}$~\cite{he2016deepresnet}, $mobilevit_{xxs}$~\cite{mehta2021mobilevit} and $mobilenet_{small0.5}$~\cite{howard2019searchingmobilenetv3}.
By default, all experiments utilized models initialized with weights pretrained on ImageNet.
Regarding the global binary decision stage, we considered the following strategies as mentioned in \Cref{subsec:proposed-decision} :\begin{itemize}
    \item \textbf{Whole video}: Decision made using all video frames. It is a greedy strategy which prevents any eventual bias related to frame selection or video duration.
    \item \textbf{Cumulative}: This strategy, utilized by \MIDVHOLO~\cite{koliaskina_midv-holo_2023}, involves iteratively updating a cumulative metric over the sequence. If the metric surpasses a predefined threshold, the video is deemed original, potentially leading to an early stop. Otherwise, if the threshold is not met by the sequence's end, the video is classified as attack.
\end{itemize}

For our weakly supervised approach, network features are trained on the train set, the best epoch is selected based on the validation set, and the decision threshold is calibrated on the validation set. Calibration involves selecting the best $F_{score}$ for both the \emph{whole video} and \emph{cumulative} decision strategies.
For the \MIDVHOLO baseline reproduction, parameters are calibrated on the union of training and validation sets.
Each operation is repeated 5 times with different train/validation/test splits, utilizing various random seeds for generation to mitigate potential biases. Results presented in Tables~\ref{tab:results} and \ref{tab:ablationstudy} represent averages and standard deviations across these 5 runs.

\subsection{Results and Ablation Study}
\label{sec:results}

\begin{table}[tb]
\centering
\caption{Comparison of results between the \MIDVHOLO baseline (reimplemented by us) and our proposed method. Metrics (F$_{score}$ or $Recall$ for attacks-only datasets) are presented across three distinct test datasets, for 2 decision strategies: Whole video and Cumulative. Both methods utilize exclusively our proposed \MIDVHOLO ``Vanilla'' train/validation sets for training and calibration. * denotes the original \MIDVHOLO configuration (applied on full rectified documents), albeit using train-validation and test splits. ``\MIDVHOLO ROI`` and our method both focus on the same ROI.
}
\label{tab:results}
\resizebox{\linewidth}{!}{%
\begin{tabular}{|c|l|l|l|l|} 
\cline{3-5}
\multicolumn{1}{c}{} & \multicolumn{1}{r|}{\textbf{Test dataset →}} & \multicolumn{1}{c|}{\begin{tabular}[c]{@{}c@{}}\textbf{MIDV-HOLO}\\\textbf{``Vanilla''}\\\textbf{(120 mixed vids)}\end{tabular}} & \multicolumn{1}{c|}{\begin{tabular}[c]{@{}c@{}}\textbf{MIDV-HOLO}\\\textbf{``Photo repl.''}\\\textbf{(20 attack vids)}\end{tabular}} & \multicolumn{1}{c|}{\begin{tabular}[c]{@{}c@{}}\textbf{MIDV 2020}\\\textbf{``Clips''}\\\textbf{(1k attack vids)}\end{tabular}} \\ 
\hline
\vcell{\textbf{Decision}} & \multicolumn{1}{r|}{\vcell{\diagbox{\textbf{Method ↓}}{\textbf{Metric →}}}} & \vcell{\textbf{$F_{score}$} (\%)} & \vcell{\textbf{$Recall$} (\%)} & \vcell{\textbf{$Recall$} (\%)} \\[-\rowheight]
\printcellbottom & \multicolumn{1}{r|}{\printcellmiddle} & \printcellmiddle & \printcellmiddle & \printcellmiddle \\ 
\hline
\multirow{2}{*}{\begin{tabular}[c]{@{}c@{}}Whole video\end{tabular}} & \MIDVHOLO ROI & 80 ± 3 & 63 ± 10 & 92 ± 2 \\
 & OUR - $mobilevit_{xxs}$ & 90 ± 2 & 87 ± 14 & 93 ± 6 \\
\hline
\multirow{3}{*}{\begin{tabular}[c]{@{}c@{}}Cumulative\end{tabular}} & \MIDVHOLO FULL DOC * & 77 ± 1 & 27 ± 12 & 84 ± 5 \\
& \MIDVHOLO ROI & 82 ± 4 & 66 ± 10 & 93 ± 0  \\
 & OUR - $mobilevit_{xxs}$ & 86 ± 5 & 84 ± 11 & 94 ± 4 \\
\hline
\multirow{3}{*}{Dummy} & Perfectly random & $50$ & $50$ & $50$ \\
 & Always positive (attack) & $67$ & $100$ & $100$ \\
 & Always negative (original) & $0$ & $0$ & $0$ \\
\hline
\end{tabular}
}
\end{table}

\Cref{tab:results} presents the outcomes achieved with the various configurations.
For brevity, only the result for the best feature extraction architecture is reported here.
The final rows of the table serve as a baseline, indicating the performance metrics for completely random and constant decision processes.
Notably, the constant prediction of attacks reaches an F$_{score}$ of 67\% on \MIDVHOLO ``Vanilla'', setting a lower bound for acceptable results.

These first results show the superiority of our proposed approach on \MIDVHOLO test sets for both decision strategies.
While achieving similar performance to the baseline on the MIDV 2020 test set, the consistently high scores may suggest a bias towards predicting attacks, contrasting with our method's robustness shown in the mixed dataset.

Finally, we conducted an ablation study to challenge the benefits of various aspects of our method.
\Cref{tab:ablationstudy} summarizes the results, with the second row being the reference for our non-ablated approach (Augmentations, Contrastive, Full train set).
The ablated components are described here below.%

\subsubsection{Data Augmentation:} \emph{What is the impact of data augmentations?} The triplet loss, along with our sampling strategy, essentially relies on augmentations.
The difference in performance between the first two rows of \Cref{tab:ablationstudy} confirms that this key component helps generalize across different datasets. Let's note that for ``Originals'' triplets, disabling augmentations nullifies the term $d(a_i, p_i)$ in \Cref{eq:loss} as $a_i$ and $p_i$ are equal.

\subsubsection{Training Strategy:} \emph{Is a contrastive loss competitive against direct decision optimization?} We trained a simple classifier under the same conditions (pretrained on ImageNet, same augmentations) to distinguish between original and attack frames. Then, the evaluation was done at the video level based on the average prediction for each frame, and the final outcome was calculated using a threshold calibrated on the validation set, similar to other methods. Results were surprisingly good on the \MIDVHOLO "Vanilla" test set but showed a significant drop on other datasets. This underscores that while \MIDVHOLO is useful as it's the first academic one of its kind, it must be handled with care. It also emphasizes the necessity of using multiple datasets to demonstrate the relevance of each method. Furthermore, it must be noted that a binary classifier cannot individually control the coverage of expected visual appearances of a hologram.

\subsubsection{Training Set:} \emph{Are attack samples required to train our approach?} The proposed method operates on the assumption that there are equal numbers of frauds and origins. However, in a real-world scenario, it is challenging to access attack samples. Thus, it makes sense to study the impact of training only on original samples. For this specific experiment, during training, the best model was selected based on a minimum validation loss criterion (over Originals only). However, the final decision threshold calibration was computed on the extended validation set (Originals and Attacks). Training only with \MIDVHOLO ``Vanilla'' Originals results in slightly lower performance on the test set, but still remains better than the \MIDVHOLO baseline.

\subsubsection{Model Architecture and Tuning:} \emph{How important is the model architecture and are pretrained weights sufficient?}
As our approach is not tied to a particular architecture, we trained and tested several lightweight ones that can match industrial processing speed requirements. We also checked whether fine-tuning actually improved performance, as pretrained weights can already exhibit sensitivity to saturated colors present in holograms. Results show similar performance for the architectures tested, with $mobilevit_{xxs}$ and $mobilenet_{small0.5}$ being slightly superior when trained on mixed samples and originals only, respectively.
Furthermore, the poor performance in the last row of \Cref{tab:ablationstudy} proves that generic features do not provide an adequate representation for our problem.

\begin{table}
\centering
\caption{Ablation study showing the contribution of 3 essential components of the proposed method: 1) data augmentation, 2) contrastive learning strategy, 3) training data. All configurations are tested on 3 different model architectures. * best configuration with all the features enabled (reported in \Cref{tab:results}).}
\label{tab:ablationstudy}
\resizebox{\linewidth}{!}{%
\begin{tabular}{|c|c|c|c|l|l|l|l|} 
\cline{6-8}
\multicolumn{4}{c}{\vcell{\begin{tabular}[b]{@{}c@{}}ABLATED ELEMENTS\\OF THE PIPELINE\end{tabular}}} & \multicolumn{1}{r|}{\vcell{\textbf{Test dataset~ →}}} & \vcell{\begin{tabular}[b]{@{}l@{}}\textbf{MIDV-Holo}\\\textbf{``Vanilla''}\\\textbf{(120 mixed vids)}\end{tabular}} & \vcell{\begin{tabular}[b]{@{}l@{}}\textbf{MIDV-Holo}\\\textbf{``Photo replace''}\\\textbf{(20 fake vids)}\end{tabular}} & \vcell{\begin{tabular}[b]{@{}l@{}}\textbf{MIDV 2020}\\\textbf{``Clips''}\\\textbf{(1k fake vids)}\end{tabular}} \\[-\rowheight]
\multicolumn{4}{c}{\printcellbottom} & \multicolumn{1}{r|}{\printcellmiddle} & \printcellmiddle & \printcellmiddle & \printcellmiddle \\ 
\hline
\vcell{\begin{tabular}[b]{@{}c@{}}\textbf{Data}\\\textbf{aug.}\end{tabular}} & \vcell{\begin{tabular}[b]{@{}c@{}}\textbf{Train}\\\textbf{strategy}\end{tabular}} & \vcell{\begin{tabular}[b]{@{}c@{}}\textbf{Training}\\\textbf{Dataset}\end{tabular}} & \vcell{\begin{tabular}[b]{@{}c@{}}\textbf{Deci-}\\\textbf{sion}\end{tabular}} & \vcell{\diagbox{\textbf{Archi }↓}{\textbf{Metric →}}} & \vcell{\textbf{\textit{F$_{score}$}} (\%)} & \vcell{\textbf{\textit{Recall}} (\%)} & \vcell{\textbf{\textit{Recall}} (\%)} \\[-\rowheight]
\printcellmiddle & \printcellmiddle & \printcellmiddle & \printcellmiddle & \printcelltop & \printcelltop & \printcelltop & \printcelltop \\ 
\hline
\multirow{3}{*}{\begin{tabular}[c]{@{}c@{}}On\end{tabular}} & \multirow{6}{*}{\begin{tabular}[c]{@{}c@{}}Contrast\\(triplet\\ loss)\end{tabular}} & \multirow{9}{*}{\begin{tabular}[c]{@{}c@{}}
MIDV-Holo\\``Vanilla''\\full train set\\(Originals \&\\ Attacks)\end{tabular}} & \multirow{15}{*}{\begin{tabular}[c]{@{}c@{}}Whole\\video\end{tabular}} & $mobilenetv3_{s50}$ & 88 ± 3 & 93 ± 8 & 92 ± 5 \\
 &  &  &  & $mobilevit_{xxs}$ * & 90 ± 2 & 87 ± 14 & 93 ± 6 \\
 &  &  &  & $resnet18$ & 88 ± 2 & 91 ± 7 & 93 ± 5 \\ 
\cline{1-1}\cline{5-8}
\multirow{3}{*}{Off} &  &  &  & $mobilenetv3_{s50}$ & 83 ± 6 & 75 ± 17 & 86 ± 7 \\
 &  &  &  & $mobilevit_{xxs}$ & 87 ± 12 & 65 ± 20 & 87 ± 7 \\
 &  &  &  & $resnet18$ & 88 ± 6 & 81 ± 13 & 83 ± 5 \\ 
\cline{1-2}\cline{5-8}
\multirow{6}{*}{\begin{tabular}[c]{@{}c@{}}\\ On\end{tabular}} & \multirow{3}{*}{\begin{tabular}[c]{@{}c@{}}Classifier\\(softmax)\end{tabular}} &  &  & $mobilenetv3_{s50}$ & 89 ± 3 & 77 ± 12 & 44 ± 7 \\
 &  &  &  & $mobilevit_{xxs}$ & 94 ± 3 & 85 ± 11 & 59 ± 4 \\
 &  &  &  & $resnet18$ & 92 ± 1 & 76 ± 10 & 76 ± 14 \\ 
\cline{2-3}\cline{5-8}
 & \multirow{3}{*}{\begin{tabular}[c]{@{}c@{}}Contrast\\(triplet\\ loss)\end{tabular}} & \multirow{3}{*}{\begin{tabular}[c]{@{}c@{}}Originals\\only\end{tabular}} &  & $mobilenetv3_{s50}$ & 82 ± 7 & 89 ± 11 & 94 ± 4 \\
 &  &  &  & $mobilevit_{xxs}$ & 84 ± 4 & 87 ± 18 & 89 ± 9 \\
 &  &  &  & $resnet18$ & 83 ± 2 & 84 ± 13 & 87 ± 8 \\ 
\cline{1-3}\cline{5-8}
\multicolumn{3}{|c|}{\multirow{3}{*}{None (pretrained weights)}} &  & \multicolumn{1}{l|}{$mobilenetv3_{s50}$} & 73 ± 6 & 81 ± 15 & 61 ± 19 \\*
\multicolumn{3}{|l|}{} &  & \multicolumn{1}{l|}{$mobilevit_{xxs}$} & 67 ± 1 & 92 ± 10 & 82 ± 7 \\*
\multicolumn{3}{|l|}{} &  & \multicolumn{1}{l|}{$resnet18$} & 77 ± 7 & 76 ± 19 & 59 ± 16 \\
 
\hline
\end{tabular}
}
\end{table}

\section{Conclusion}
We have presented a novel approach for verifying both Appearance Conformity and Appearance Coverage of Optically Variable Devices (OVDs, or ``holograms'') in identity documents using video clips recorded from commodity smartphones. This approach leverages a feature extraction network trained with a contrastive loss, which can be specialized to a given hologram while requiring only video-level annotations, rather than individual frame labels. Furthermore, we have demonstrated that this approach can achieve attractive results using original video samples alone, which are abundantly obtained in industrial pipelines. Thanks to the separate calibration of its decision stage, our approach can be easily tuned to specific security requirements.

The evaluation of this approach necessitated the introduction of several extensions to the original \MIDVHOLO dataset and the reimplementation of the proposed baseline. Our experiments have revealed the superiority of our approach over the previous baseline and its robust generalization capabilities across both the MIDV-Holo ``Photo Replacement'' and MIDV 2020 ``Clips''.

Lastly, our ablation study has uncovered a significant insight: while the MIDV-Holo ``Vanilla'' dataset yields intriguingly good results when tested with a simple binary classifier trained at the frame level, its generalization to other datasets is poor, as expected. This raises the question: \emph{``What does the binary classifier actually learn?''} for future investigation.

\printbibliography

\subsection*{Acknowledgements}
The SOTERIA project, partially supporting this work, was funded by the European Union's Horizon 2020 research and innovation program under grant agreement No 101018342.

\justifying
\end{document}